%% file: main_cvpr.tex
\documentclass[10pt,twocolumn,letterpaper]{article}

 \usepackage{cvpr}              

\input{preamble}

\definecolor{cvprblue}{rgb}{0.21,0.49,0.74}
\usepackage[pagebackref,breaklinks,colorlinks,citecolor=cvprblue]{hyperref}

\def\paperID{11544} 
\def\confName{CVPR}
\def\confYear{2024}

\usepackage{times}  
\usepackage{caption}
\usepackage{subcaption}

\usepackage{helvet}  
\usepackage{courier}  
\usepackage[hyphens]{url}  
\usepackage{graphicx} 
\usepackage{natbib}  
\usepackage{algorithm}
\usepackage[justification=centering]{caption}
\usepackage{booktabs}       

\usepackage{algorithmic}

\floatstyle{ruled}
\newfloat{listing}{tb}{lst}{}
\floatname{listing}{Listing}
\usepackage{bibentry}
\usepackage{xspace}
\makeatletter
\DeclareRobustCommand\onedot{\futurelet\@let@token\@onedot}
\def\@onedot{\ifx\@let@token.\else.\null\fi\xspace}

\def\ie{\emph{i.e}\onedot} 
 
\def\etc{\emph{etc}\onedot}

\def\etal{\emph{et al}\onedot}

\makeatother

\usepackage{listings}

\usepackage{xcolor}

\definecolor{citecolor}{RGB}{34,139,34}
\definecolor{codegreen}{rgb}{0,0.5,0}
\definecolor{codeblue}{rgb}{0.25,0.5,0.5}
\definecolor{codegray}{rgb}{0.6,0.6,0.6}

\title{BCN: Batch Channel Normalization for Image Classification}

\author{Afifa Khaled \\
{\tt\small afifakhaied@tju.edu.cn}
\and
Chao Li \\
{\tt\small D201880880@hust.edu.cn}
\and
Jia Ning \\ 
{\tt\small ninja@hust.edu.cn}
\and
Kun He \\
{\tt\small brooklet60@hust.edu.cn}\\
School of Computer Science and Technology, Huazhong University of Science and Technology,\\ Wuhan 430074, China\\
}
\begin{document}
\maketitle
\input{sec/0_abstract}    
\input{sec/1_intro}

\input{sec/2_RW}

\input{sec/3_Method}

\input{sec/4_ex}

\input{sec/5_con}
{
   \small
    \bibliographystyle{ieeenat_fullname}
    \bibliography{cvpr2024_conference}
}
\end{document}

%% file: preamble.tex
\usepackage[dvipsnames]{xcolor}


%% file: sec/0_abstract.tex
\begin{abstract}
Normalization techniques have been widely used in the field of deep learning due to their capability of enabling higher learning rates and are less careful in initialization. However, the effectiveness of popular normalization technologies is typically limited to specific areas. 
Unlike the standard Batch Normalization (BN) and Layer Normalization (LN), where BN computes the mean and variance along the (N,H,W) dimensions and LN computes the mean and variance along the (C,H,W) dimensions (N, C, H and W are the batch, channel, spatial height and width dimension, respectively), this paper presents a novel normalization technique called Batch Channel Normalization (BCN). To exploit both the channel and batch dependence and adaptively and combine the advantages of BN and LN based on specific datasets or tasks, 
BCN separately normalizes inputs along the (N, H, W) and (C, H, W) axes, then combines the normalized outputs based on adaptive parameters. 
As a basic block, BCN can be easily integrated into existing models for various applications in the field of computer vision. Empirical results show that the proposed technique can be seamlessly applied to various versions of CNN or Vision Transformer architecture. The code is publicly available at https://github.com/AfifaKhaled/Batch-Channel-Normalization.
\end{abstract}

%% file: sec/1_intro.tex
\section{Introduction}
\label{sec:intro}
In the past decades, machine learning (ML) 
has become the most widely used technique in the field of artificial intelligence, 
and more recently, deep learning (DL) has become a prevalent topic, and deep neural networks (DNNs) are widely applied in various domains, including natural language processing, computer vision, and graph mining. Typically, DNNs comprise stacked layers with learnable parameters and non-linear activation functions. While the deep and complex structure enables them to learn intricate features, it also poses challenges during training due to the randomness in parameter initialization and changes in input data, known as internal covariate shift \citep{b1}. This problem becomes more pronounced in deeper networks, where slight modifications in deeper hidden layers are amplified as they propagate through the network, resulting in significant shifts in these layers.

To address the above issue, several normalization methods have been introduced.  
Specifically, Batch Normalization (BN)~\citep{b1}, Layer Normalization (LN)~\citep{b2}, and Group Normalization (GN)~\citep{b3} \etal
have achieved remarkable success with deep learning models. Among them, BN is widely used for deep neural networks.

Despite the great success in many applications, the popular normalization methods still have some weaknesses. For example, BN
requires large batch sizes \citep{b3},  can not be used for online learning tasks, and can not be used for large distributed models because the mini-batches have to be small. To address these issues, LN is proposed to avoid exploiting batch dimension so as to not impose any restriction on the size of each mini-batch \citep{b2}. However, LN does not work as well as BN on convolutional layers.





\begin{figure*}[t]
\begin{center}  
\includegraphics[width=1.0\textwidth]{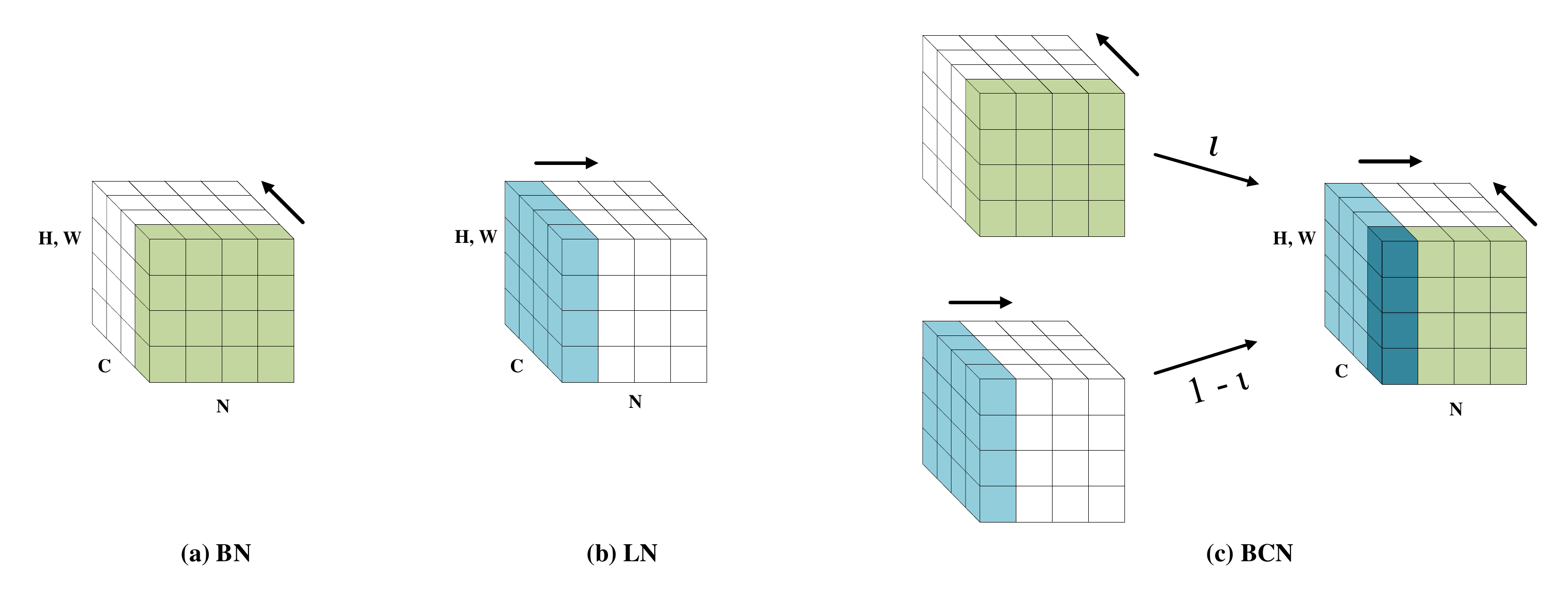}
\caption{Visualization on several normalization techniques. Each subplot shows a feature map tensor with N the batch axes, C the channel axes, and (H, W) the spatial height and width axes.
}
\label{fig}
\end{center}
\end{figure*}

To overcome the limitations of BN and LN as well as to fully embody the advantages of the two techniques, we develop a new normalization technique called Batch Channel Normalization (BCN). In contrast to previous techniques, 
we aim to normalize along the (C, N, H, W) axes. However, computing the average and variance along (N, C, H, W) directly ignores the different importance between batch dimension and channel dimension. Consequently, as shown in Figure~\ref{fig},
BCN first computes the $\mu_1$ and $\sigma^2_1$ of the layer inputs along the (C, H, W) axes. Then, it computes the $\mu_2$ and $\sigma^2_2$ along the (L, H, W) axes. Finally, the normalized outputs are combined based on adaptive parameters. To check the effectiveness of the proposed method, we apply BCN to several popular models, ResNet~\citep{b5}, DenseNet~\citep{B22},  Vision Transformer ~\citep{b44} and BYOL~\citep{b10}, on the image classification task.  
Our experiments demonstrate that BCN yields promising results, leading to improved training speed and enhanced generalization performance.

Our main contributions are summarized as follows:

\begin{itemize}
    \item 
    We introduce a new normalization technique termed Batch Channel normalization (BCN) as a simple alternative to BN and LN techniques.  
    \item BCN exploits the channel and batch dependence, and adaptively combines the information of channel dimension and batch dimension, which embodies the advantages of both BN and LN.
    \item Empirically, we show that our BCN normalization technique can substantially improve the generalization performance of the neural networks compared to existing normalization techniques.
\end{itemize}

%% file: sec/2_RW.tex
\section{Related Work}
\subsection{Dimension Normalization} 
The first group involves normalizing different dimensions of the output. Examples include Layer Normalization \citep{b2}, which normalizes inputs across features; Instance Normalization \citep{b4}, which normalizes over spatial locations in the output; and Group Normalization \citep{b3}, which independently normalizes along spatial dimensions and feature groups. 

Here we introduce two that are most related to our work, \ie, Batch Normalization~\citep{b1} and Layer Normalization \citep{b2}.

\textbf{Batch Normalization (BN)} allows faster convergence and stabilizes the learning. During the training set, BN computes the mean  $\mu_B$ and variance  $\sigma^2_B$ of the layer inputs as follows:

\begin{equation}
\mu_B = \frac{\displaystyle 1}{n} {\sum_{i=1}^{n} x_i},
\end{equation}

\begin{equation}
 \sigma^2_B =  \frac{\displaystyle 1}{n}\sum_{i=1}^{n}(x_i - \mu_B)^2,
\end{equation}

\begin{equation}
 \bar x_B = \gamma  \frac{\displaystyle (x_i - \mu_B)}  {\sqrt { (\sigma^2_B +\epsilon)}} +  \beta.
\end{equation}

We can see that BN computes the $\mu_B$ and $\sigma^2_B$ along the (N, H, W) axes \citep{b1}. 
During the testing, BN computes the $\mu_B$ and $\sigma^2_B$ by exponential moving average during the training set: 
\begin{equation}
\mu = \alpha \mu + (1- \alpha) \mu_B, 
\end{equation}

\begin{equation}
\bar x = \gamma  \frac{\displaystyle (x_i - \mu)}  {\sqrt { (\sigma^2 +\epsilon)}} +  \beta,
\end{equation}
where n is batch size, $ \gamma $ and $ \beta $ are learnable parameters.
Here $\alpha$ is usually set to 0.9, and $\epsilon$ is a small constant.

\textbf{Layer Normalization (LN)} computes the  $\mu_L$ and $\sigma^2_L$  along the (C, H, W) as follows:
\begin{equation}
\mu_L = \frac{\displaystyle 1}{n} {\sum_{i=1}^{n} x_i}, 
\end{equation}
\begin{equation}
\sigma^2_L =  \frac{\displaystyle 1}{n}\sum_{i=1}^{n}(x_i - \mu)^2, 
\end{equation}
\begin{equation}
\bar x_L = \gamma  \frac{\displaystyle (x_i - \mu_L)}  {\sqrt { (\sigma^2_L +\epsilon)}} +  \beta.
\end{equation}
Unlike BN, LN performs the same computation at training and inference times. Moreover, LN is very useful for stabilizing the dynamics of hidden states in recurrent neural networks.

Our method belongs to this group. The proposed BCN aims to normalize along to (C, N, H, W) axes, which aims to combine the benefits of both BN and LN while mitigating their respective deficiency.

\subsection{Normalization Improvement}
The second group modifies the original batch normalization method \citep{b1}. This group includes methods like Ghost BN \citep{hoffer2017train}, which normalizes independently across different splits of batches, and Batch Re-normalization \citep{ioffe2017batch} or Streaming Normalization \citep{liao2016streaming}, both of which make changes to utilize global averaged statistics instead of batch statistics. 

While these normalization techniques are practically famous and successful, their improvement has only started to appear recently. At the same time, Batch normalization \citep{b1} remains the most famous normalization technique used so far. In addition, these normalization techniques have not been able to approach BN’s accuracy in many tasks (e.g., segmentation, detection, and video classification).

\subsection{Weight Normalization}
The third group consists of methods that normalize weights rather than activations. This group comprises Weight Normalization \citep{salimans2016weight} and Normalization Propagation \citep{arpit2016normalization}, both of which divide weights by their $\ell_2$ norm, differing only in minor details.

Several papers have recently proposed techniques to enhance the weight normalization across a wider range of CNN. One approach is to implicit regularization and convergence for weight normalization \citep{b43}. They studied the weight normalization technique and reparametrized projected gradient descent for over-parameterized least squares regression. They showed 
that the non-convex formulation has useful regularization effects compared to gradient descent on the original objective.          Due to the limited number of pages, we refer the reader to read more normalization techniques at https://github.com/AfifaKhaled/Normalization-Techniques--survey.





%% file: sec/3_Method.tex
\section{Methodology}


The motivation behind the success of the normalization techniques has been an important research topic. In this section, we investigate the motivation for developing the new normalization technique. Our research has revealed that the normalization goal for all normalization techniques is to improve the model's robustness \citep{b9}.



\subsection{Method Formulation}
The idea of normalizing the (N, H, W) axes and (C, H, W) axes has been proposed before \citep{b1,b2,b4,b3}. Earlier works typically perform normalization along (N, H, W) axes or (C, H, W) axes independently. We aim to perform normalization along (N, C, H, W). However, computing the average and variance along (N, C, H, W) directly ignores the different importance between batch dimension and channel dimension. Consequently, we propose to separately normalize along the (N, H, W) and (C, H, W) axes, then combine the normalized outputs based on adaptive parameters $\iota$. 
Doing so could improve the training, validation, and test accuracy, as we show experimentally in the next section. 

In a similar fashion to how BN normalizes the layer inputs, during the training, BCN first computes the average $\mu_1$ and the variance $ \sigma^2_1$ of the layer inputs along (N, H, W ) axes:
\begin{equation}
\mu_1 = \frac{\displaystyle 1}{n} {\sum_{i=1}^{n} x_i},
\label{1}
\end{equation}
\begin{equation}
\sigma^2_1 =  \frac{\displaystyle 1}{n}\sum_{i=1}^{n}(x_i - \mu_1)^2.
\label{2}
\end{equation}


Second, BCN computes the average $\mu_2$ and variance $ \sigma^2_2$ along (C, H, W ):
\begin{equation}
\mu_2 = \frac{\displaystyle 1}{n} {\sum_{i=1}^{n} x_i},
\label{3}
\end{equation}
\begin{equation}
\sigma^2_2 =  \frac{\displaystyle 1}{n}\sum_{i=1}^{n}(x_i - \mu_2)^2.
\label{4}
\end{equation}


Next, $\bar x_1$ and $\bar x_2$ are normalized using $\mu_1$, $ \sigma^2_1$ and $\mu_2$, $ \sigma^2_2$, respectively:
\begin{equation}
\bar x_1= \frac{\displaystyle (x_i - \mu_1)}  {\sqrt { (\sigma^2_1 +\epsilon)}},
 \label{5}
\end{equation}
\begin{equation}
    \bar x_2=  \frac{\displaystyle (x_i - \mu_2)}  {\sqrt { (\sigma^2_2 +\epsilon)}}.  \;                         
    \label{6}
\end{equation}

BCN introduces additional learnable parameter $\iota$ to adaptively balance the normalized outputs along the axes of (N, H, W)  and (C, H, W).
\begin{equation}
    \bar y = \iota \bar x_1 + (1- \iota)  \bar x_2,
    \label{7}
\end{equation}
Then, the output of BCN normalization can be formulated as follows:
\begin{equation}
   Y= \gamma  \bar y + \beta , \;        
   \label{8}
\end{equation}
\noindent where $ \gamma $ and $ \beta $ are learnable parameters and $\epsilon$ is a small constant for numerically stability.


At the inference stage, since the $\mu$ and the $\sigma$ are pre-computed and fixed, the normalization can be fused into the convolution operation.

Following previous works~\citep{cai2021exponential, luo2020extended}, 
BCN normalizes along the (N, H, W) axes by exponential moving average \citep{b14} during the training as follows: 
\begin{equation}
\mu = \alpha  \mu + (1- \alpha) \mu_1, 
\end{equation}
\begin{equation}
\sigma^2 = \alpha   \sigma^2 + (1- \alpha)  \sigma^2_1,
\end{equation}
\begin{equation}
\bar x =  \frac{\displaystyle (x_i - \mu)}  {\sqrt  {(\sigma^2 +\epsilon)}},
\end{equation}
\noindent where $\alpha$ is set to 0.9 in our experiments.

The key difference between BCN normalization and existing normalization techniques is that under BCN, all the channels in a layer share the same normalization terms $\mu$ and $ \sigma^2$.

\subsection{Implementation}
BCN can be implemented by a few lines of Python code in PyTorch~\citep{b8} or TensorFlow~\citep{b7} where computing $\bar x_1$  along  (N, H, W) and $\bar x_2$ along (C, H, W) is implemented. The overall BCN process is presented in Algorithm \ref{alg:BCN}, with the corresponding Python code in Figure \ref{fig:code}.

%% file: sec/4_ex.tex
\section{Experiments and Discussion}
\begin{algorithm}[t]
\caption{Batch Channel Normalization (BCN)}
\label{alg:BCN}
\begin{algorithmic}[1]
\REQUIRE~~\\
Input $x =\{x_1, x_2, ..., x_n\}$, Parameters to be learned: $\iota$, $\beta$ and $\gamma$\\
\ENSURE ~~\\
$Y= BCN_{\gamma, \beta, \iota} (x_i)$\\    
\STATE Calculate  $\mu_1$ and $ \sigma^2_1$ based on Eq.~\ref{1} and ~\ref{2}\\
\STATE Calculate  $\mu_2$ and $ \sigma^2_2$ based on Eq.~\ref{3} and \ref{4}\\
\STATE Calculate the normalized output $ \bar x_1$ along (N, H, W) and $\bar x_2$ along (C, H, W) axes by Eq.~\ref{5} and ~\ref{6}\\
\STATE Adaptively combine $ \bar x_1$ and $ \bar x_2$ based on Eq.~\ref{7}\\
\STATE Calculate the final output Y based on Eq.~\ref{8}
\RETURN Y
\end{algorithmic}
\end{algorithm}

\lstset{
  backgroundcolor=\color{white},
  basicstyle=\fontsize{7.5pt}{8.5pt}\fontfamily{lmtt}\selectfont,
  columns=fullflexible,
  breaklines=true,
  captionpos=b,
  commentstyle=\fontsize{8pt}{9pt}\color{codegray},
  keywordstyle=\fontsize{8pt}{9pt}\color{codegreen},
  stringstyle=\fontsize{8pt}{9pt}\color{codeblue},
  frame=tb,
  otherkeywords = {self},
}
\begin{figure*}[!t]
\tiny
\begin{lstlisting}[language=python]
def BatchChannelNorm(x, gamma, beta, momentum=0.9, num_channels, eps=1e-5):
    self.num_channels = num_channels
    self.epsilon = epsilon
    self.x1 = BCN_1(self.num_channels, epsilon=self.epsilon)    # normalized along (N, H, W) axes.
    self.x2 = BCN_2(self.num_channels, epsilon=self.epsilon)     # normalized along (C, H, W) axes.
    # x: input features with shape [N,C,H,W]
    # gamma, beta: scale and offset
    self.gamma = nn.Parameter(torch.ones(num_channels))
    self.beta = nn.Parameter(torch.zeros(num_channels))
    # iota is the BCN variable to be learned to adaptively balance the normalized outputs along (N, H, W) axes and (C, H, W) axes.
    self.iota = nn.Parameter(torch.ones(self.num_channels))
    X = self.x1(x)
    Y = self.x2(x)
    Result = self.iota.view([1, self.num_channels, 1, 1]) * X + ( 1 - self.iota.view([1, self.num_channels, 1, 1])) * Y
    Result = self.gamma.view([1, self.num_channels, 1, 1]) * Result + self.beta.view([1, self.num_channels, 1, 1])  
    return Result
\end{lstlisting}
\caption{Python code of Batch Channel normalization (BCN) based on PyTorch.}
\label{fig:code}
\vspace{-1em}
\end{figure*}


\subsection{Datasets}
We evaluate the effectiveness of our technique through four representative datasets: CIFAR-10/100~\citep{b6}, SVHN~\citep{B20}, and ImageNet~\citep{B21}.
The CIFAR-10/100 datasets, developed by the Canadian Institute for Advanced Research, are widely employed in various experiments. CIFAR-10 consists of 60,000 32x32 color images divided into 10 object classes, with 50,000 training images and 10,000 test images. On the other hand, CIFAR-100 comprises 100 classes with 600 images per class \citep{b6}. 
The Street View House Numbers (SVHN) dataset \citep{B20} contains 600,000 32×32 $RGB$ images of printed digits (from 0 to 9) cropped from pictures of house number plates. 
ImageNet dataset \citep{B21} has 1.28M training images and 50,000 validation images with 1000 classes.

\subsection{Experimental Setup}
To investigate how BCN and the existing normalization
techniques work, we conduct a series of experiments.  Five normalization techniques, \ie, 
BN~\citep{b1}, 
LN~\citep{b2}, 
IN~\citep{b4}, 
GN~\citep{b3}, 
and our BCN, 
are implemented from scratch in Pytorch \citep{b8}.

The experimental details are the same in the five techniques  (\ie, loss function,  batch size, \etc). On ImageNet, we evaluate the performance of ResNet18, VGG16, SqueezeNet and AlexNet with BN, LN and BCN, respectively. 
We use accuracy on different datasets to investigate the effectiveness of the BCN normalization technique.

\begin{figure}[!t]
 \centering
 \begin{subfigure}[CIFAR-10]{0.30\textwidth}
   {\includegraphics[scale=0.5]{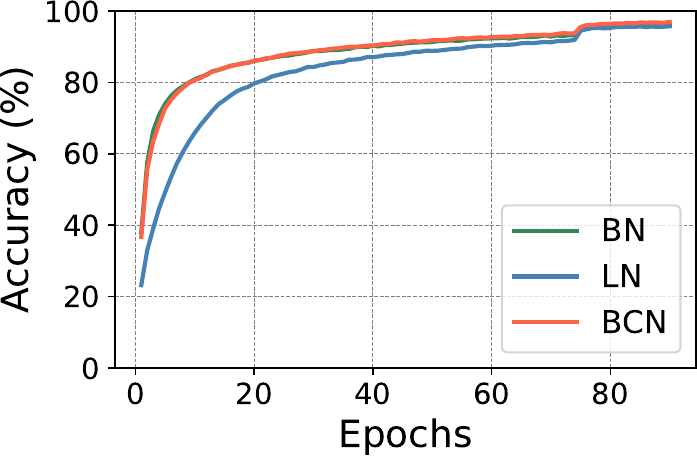}}
      \captionsetup{justification=centering}

      \caption{CIFAR-10}
   \end{subfigure}
\hspace{0.5em}
 \begin{subfigure}[CIFAR-100]{0.30\textwidth}{
   \includegraphics[scale=0.5]{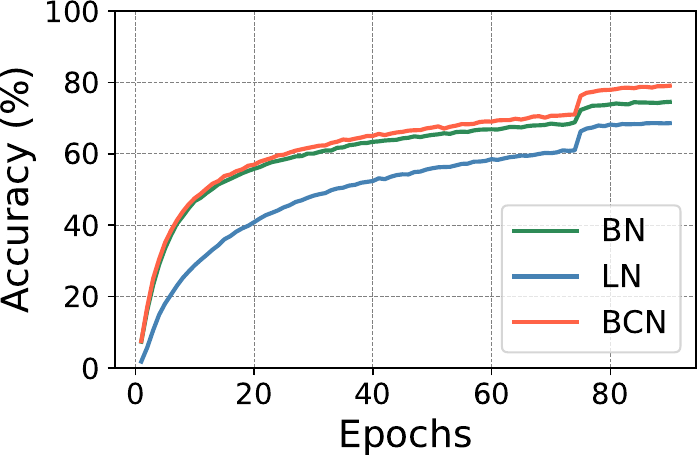}}
      \captionsetup{justification=centering}

      \caption{CIFAR-100}
  \end{subfigure}
\hspace{0.5em}
 \begin{subfigure}[SVHN]{0.30\textwidth}{
   \includegraphics[scale=0.5]{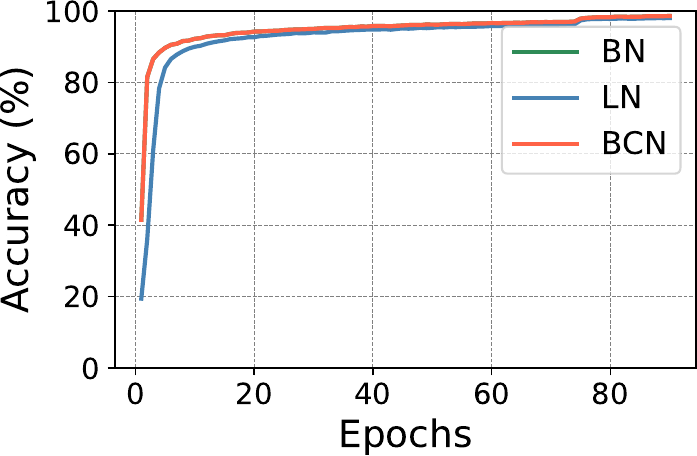}}
      \captionsetup{justification=centering}

      \caption{SVHN}
   \end{subfigure}
   \caption{Training accuracy of ResNet with different normalization techniques on (a) CIFAR-10, (b) CIFAR-100, (c) SVHN.}
 \label{fig1}
\end{figure}

\begin{figure}[!t]
 \centering
 \begin{subfigure}[CIFAR-10]{0.30\textwidth}
   {\includegraphics[scale=0.5]{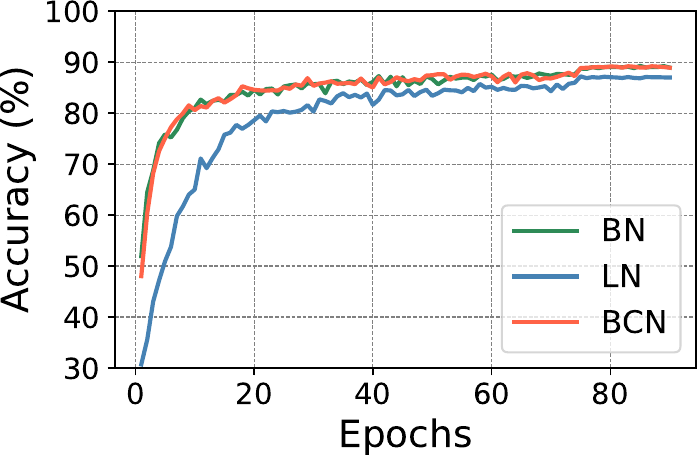}}
      \captionsetup{justification=centering}

      \caption{CIFAR-10}
   \end{subfigure}
\hspace{0.5em}
 \begin{subfigure}[CIFAR-100]{0.30\textwidth}{
   \includegraphics[scale=0.5]{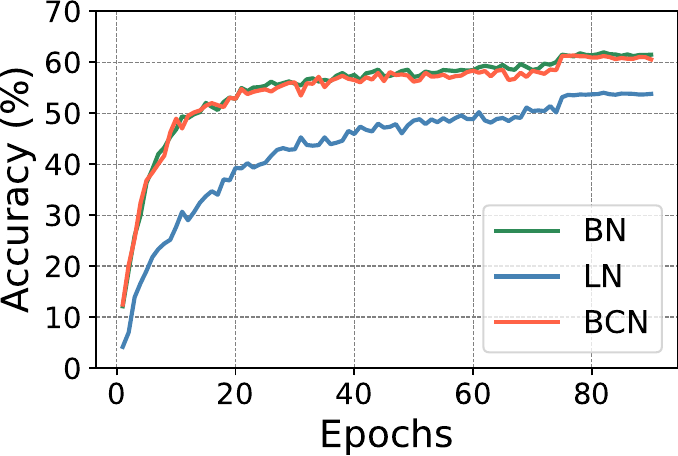}}
      \captionsetup{justification=centering}

      \caption{CIFAR-100}
  \end{subfigure}
\hspace{0.5em}
 \begin{subfigure}[SVHN]{0.30\textwidth}{
   \includegraphics[scale=0.5]{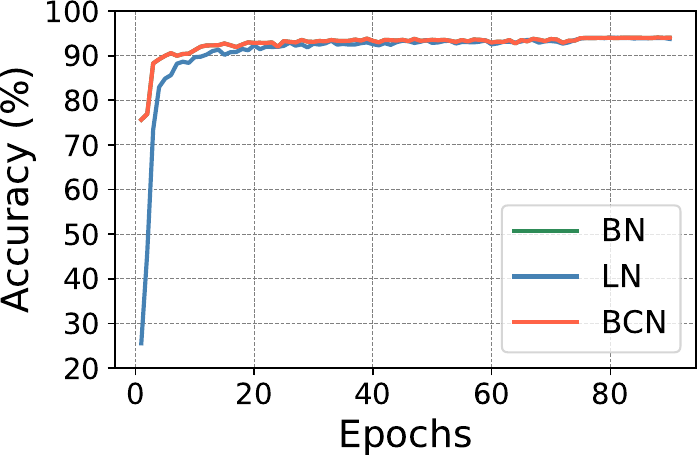}}
      \captionsetup{justification=centering}

      \caption{SVHN}
   \end{subfigure}
   \caption{Validation accuracy of ResNet with  different normalization techniques on (a) CIFAR-10, (b) CIFAR-100, (c) SVHN.}
 \label{fig2}
\end{figure}
 



\begin{figure}[!t]
 \centering
 \begin{subfigure}[Train]{0.30\textwidth}
   {\includegraphics[scale=0.5]{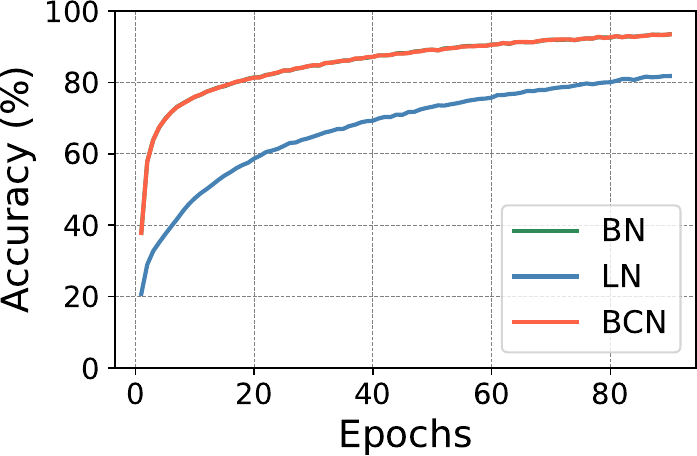}}
      \captionsetup{justification=centering}
      \caption{Train}
   \end{subfigure}
 \begin{subfigure}[Validation]{0.30\textwidth}{
   \includegraphics[scale=0.5]{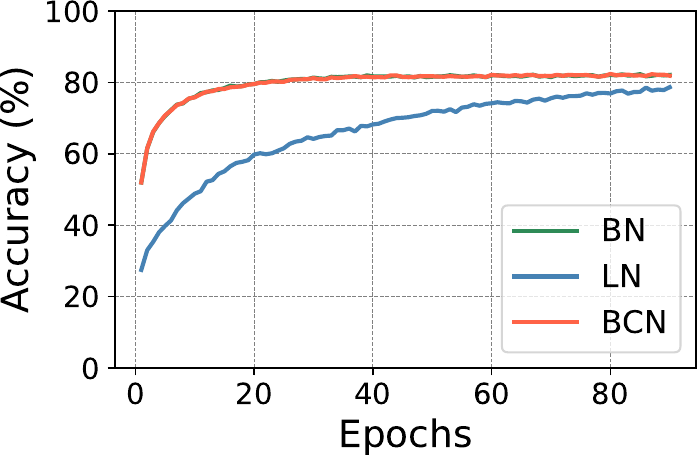}}
      \captionsetup{justification=centering}
      \caption{Validation}
  \end{subfigure}
\hspace{0.5em}
   \caption{Training and validation accuracy curve of different normalization techniques for BYOL on the CIFAR-10 dataset.
Note that the results of BN are exactly the same as that of BCN.}
 \label{fig3}
\end{figure}

 \begin{figure}[!t]
  \centering
  \begin{subfigure}[CIFAR-10]{0.30\textwidth}
    {\includegraphics[scale=0.5]{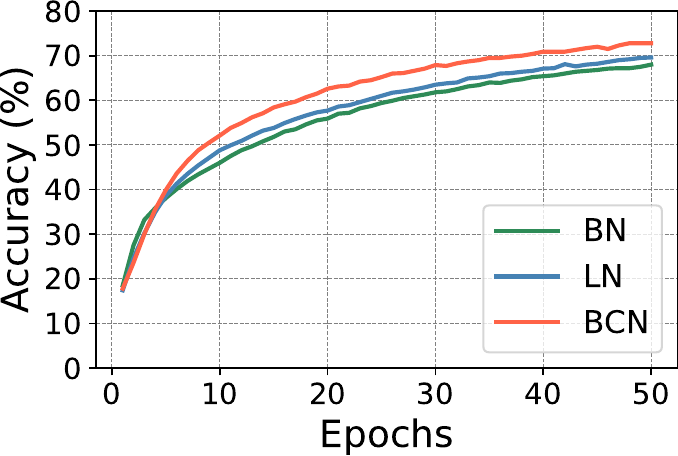}}
       \captionsetup{justification=centering}
       \caption{CIFAR-10}
    \end{subfigure}
 \begin{subfigure}[CIFAR-100]{0.30\textwidth}{
    \includegraphics[scale=0.5]{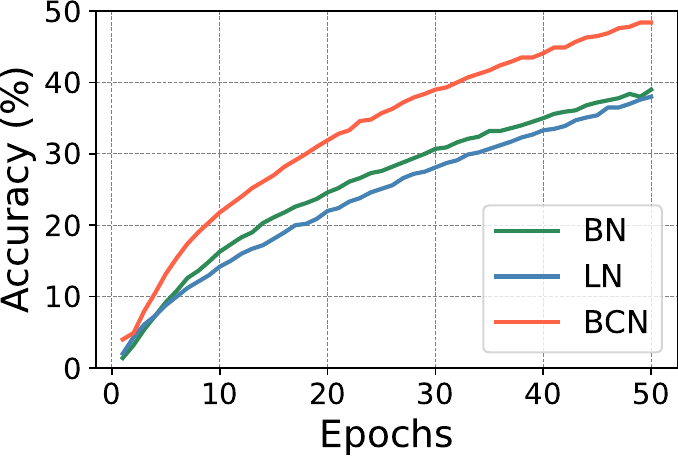}}
      \captionsetup{justification=centering}
       \caption{CIFAR-100}
   \end{subfigure}
    \caption{Training accuracy of ViT with  different normalization techniques on (a) CIFAR-10, (b) CIFAR-100.}
  \label{fig20}
 \end{figure}

 \begin{figure}[!t]
  \centering
  \begin{subfigure}[CIFAR-10]{0.30\textwidth}
    {\includegraphics[scale=0.5]{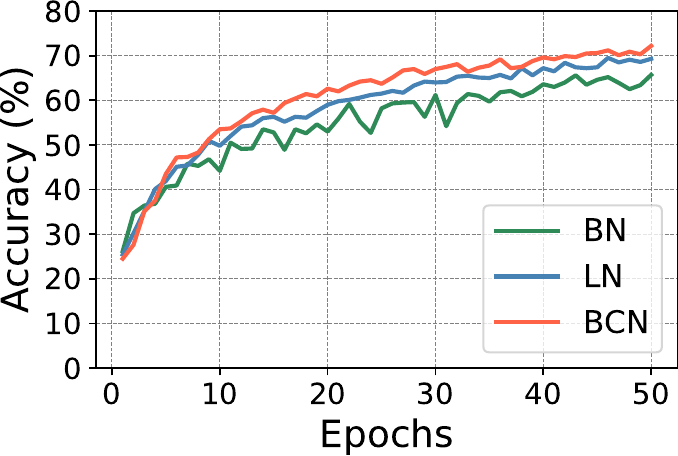}}
       \captionsetup{justification=centering}
       \caption{CIFAR-10}
    \end{subfigure}
\hspace{0.5em}
  \begin{subfigure}[CIFAR-100]{0.30\textwidth}{
    \includegraphics[scale=0.5]{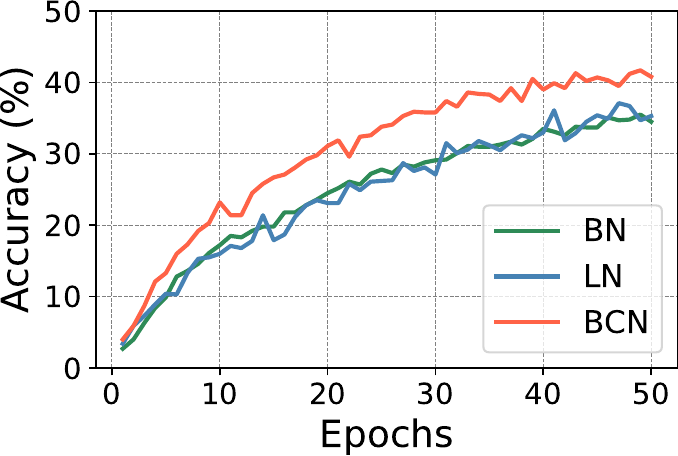}}
      \captionsetup{justification=centering}

       \caption{CIFAR-100}
   \end{subfigure}

    \caption{Testing accuracy of ViT with  different normalization techniques on (a) CIFAR-10, (b) CIFAR-100.}
  \label{fig21}
 \end{figure}

\begin{figure}[!t]
\centering
\begin{subfigure}[2-batch size]{0.28\textwidth}{
\centering
\includegraphics[scale=0.5]{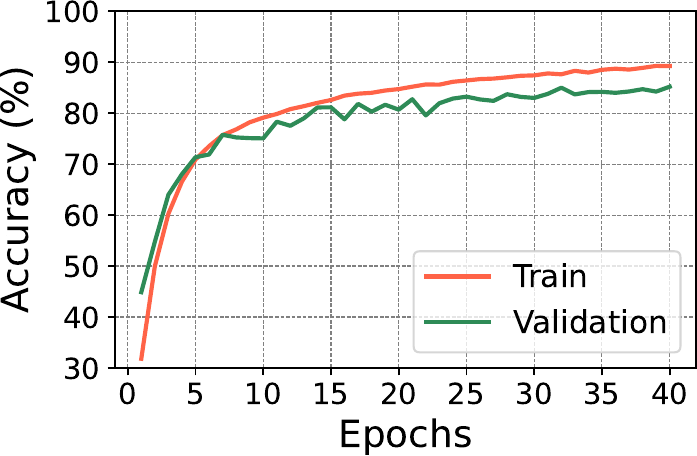}}
   \captionsetup{justification=centering}
   \caption{2-batch size}
\end{subfigure}
\begin{subfigure}[4-batch size]{0.28\textwidth}{
\centering
\includegraphics[scale=0.5]{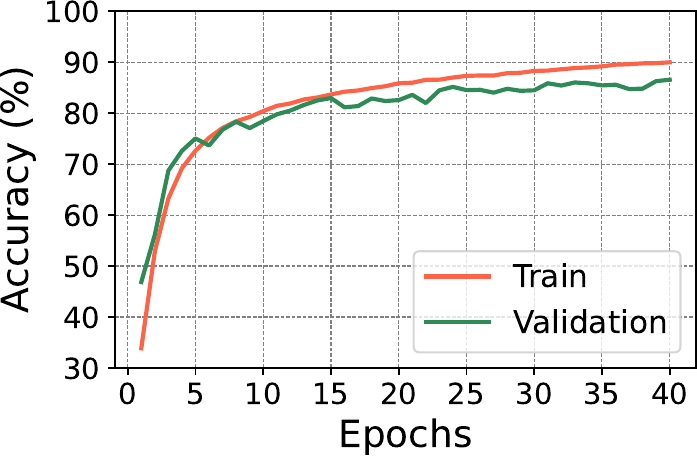}}
   \captionsetup{justification=centering}
\caption{4-batch size}
\end{subfigure}
\caption{The accuracy curve for mini-batch size on the CIFAR-10 dataset.}
\label{fig4}
\end{figure}

\begin{figure}[!t]
\centering
\begin{subfigure}[128]{0.30\textwidth}{
\includegraphics[scale=0.5]{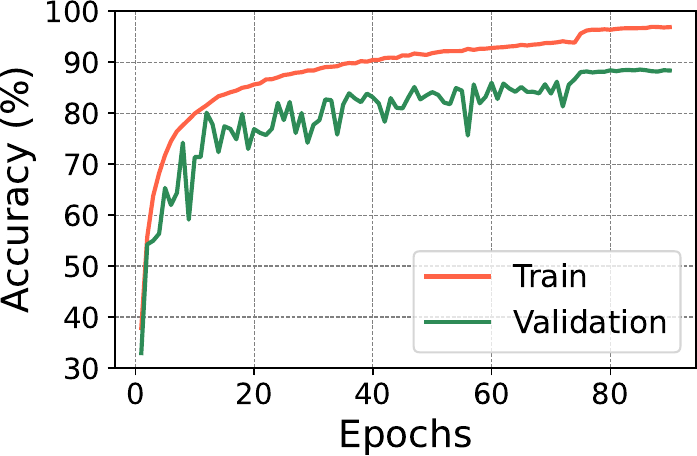}} 
   \captionsetup{justification=centering}

\caption{128-batch size}
\end{subfigure}
\hspace{0.5em}
\begin{subfigure}[16]{0.30\textwidth}{
\includegraphics[scale=0.5]{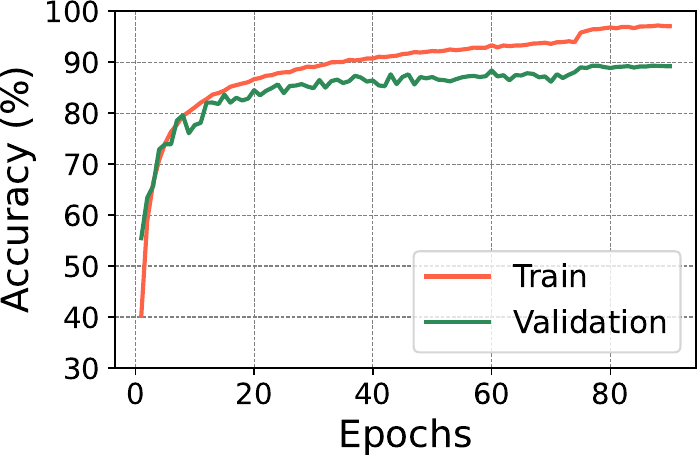}}
  \caption{16-batch size}
\end{subfigure}
\hspace{0.5em}
\begin{subfigure}[8]{0.30\textwidth}{
\includegraphics[scale=0.5]{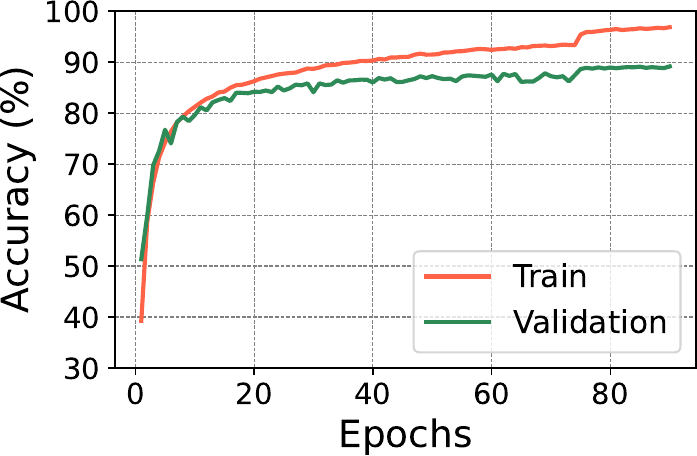}}
  \caption{8-batch size}
\end{subfigure}
\caption{The training and validation accuracy curve for different batch sizes on the CIFAR-10 dataset.}
\label{fig5}
\end{figure}

\subsection{Comparison with Normalization Techniques}
In this subsection, we compare our method with typical normalization techniques using popular datasets and neural networks. Specifically, we compare the image classification performance of ResNet with different normalization techniques on CIFAR-10/100 and SVHN datasets. We also compare different normalization techniques on DenseNet using the ImageNet dataset and self-supervised learning on BYOL using the CIFAR-10 dataset. In addition, exploring BCN on new models like Vision Transformers using the CIFAR-10/100 and SVHN datasets.

\subsubsection{Results on ResNet}
We perform experiments on ResNet~\citep{b5} for the image classification task. The model is trained by stochastic gradient descent (SGD) starting with a learning rate of 0.1 and then reduced by a factor of 10 at the 75th and 85th epochs, respectively. A batch size of 8 and a momentum of 0.9 are used.

\begin{table*}
\centering
\begin{tabular}{lccc}
        \toprule
           \textbf{Method\qquad}& \textbf{\quad~~CIFAR-10~~\quad}       & \textbf{\quad~~~CIFAR-100~~~\quad}         & \textbf{\quad~~~SVHN~~~\quad} \\
           \midrule
            BN & 96.11 &\underline{74.50}  &    98.22   \\
            LN & 95.76 & 68.61  &   97.62  \\
            IN & \underline{96.68} &  73.42  & \bfseries 98.93  \\
            GN & 95.91& 70.15 & 98.49  \\
            \bfseries BCN&\bfseries96.97 &\bfseries 79.09&\underline{98.63} \\
            \bottomrule
\end{tabular}
\caption{ 
Comparison of the test accuracy on three datasets. The best results appear in bold, and the second best are underlined. }
\label{Res_1}
\end{table*}

We show results of BCN, BN, and LN during training and validation on the three datasets in Figure~\ref{fig1} and Figure~\ref{fig2}. As we can see, BCN learns most rapidly. On CIFAR-10, in about 20 epochs, it has achieved about 86.12\% training accuracy and 84.16\% validation accuracy, whereas in the same number of epochs, the BN and LN show 86.04\% and 79.74\% training accuracy and 12.87\% and 78.58\% validation accuracy, respectively.

In addition, Table \ref{Res_1} shows the results of test accuracy for BCN normalization and the representative normalization techniques (BN , LN,  IN , and  GN ) on the CIFAR-10, CIFAR-100, and SVHN datasets. All experiments are conducted under the same learning rate, loss function, batch size, \etc 
The results 
show that BCN is generally applicable, gaining the best or the second best results. 
For example, under the CIFAR-100 dataset, BCN achieves significant improvement over the state-of-the-art techniques. 

\subsubsection{Results on BYOL}
We apply BCN to a recent state-of-the-art method BYOL~\citep{b10} for self-supervised learning. We have implemented BYOL in PyTorch, using hyperparameter settings as in the original paper \citep{b10}. 
BCN is applied in both online and target models. As shown in Figure \ref{fig3}, applying BCN can improve the performance of BYOL. 


\subsubsection{Results on ViT}
Nowadays, there is growing interest in developing Vision Transformer (ViT) methods \citep{b44} across a wider range of applications. We implemented ViT from scratch. We have tested ViT having different batch sizes and embedding dimensions on CIFAR-10/100 and SVHN datasets. Figure  \ref{fig20} shows that the performance of ViT improved when the normalization technique was replaced by BCN.  Replacing BCN lead to nearly 0.73 \%  training accuracy and 0.72 \% testing accuracy. 
In addition, Figure  \ref{fig21}  shows the performance of BCN in the testing set.
Overall, these results support our hypothesis that compared to the existing normalization techniques BN and LN, BCN can have a better performance in new models like ViT. 

\begin{figure}
\hspace{0.5em}
 \begin{subfigure}[LN]{0.30\textwidth}{
   \includegraphics[scale=0.5]{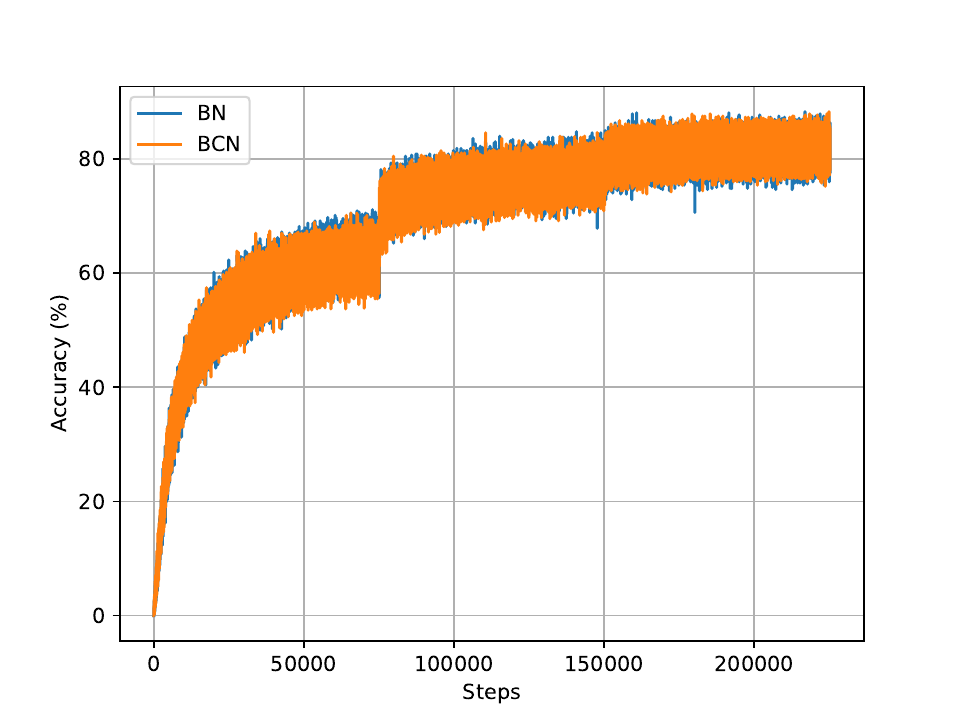}}
      \captionsetup{justification=centering}
   \end{subfigure}
   \caption{Training accuracy of DenseNet with  ImageNet dataset.}
 \label{fig11}
\end{figure}

\begin{figure}
 \begin{subfigure}[LN]{0.30\textwidth}{
   \includegraphics[scale=0.5]{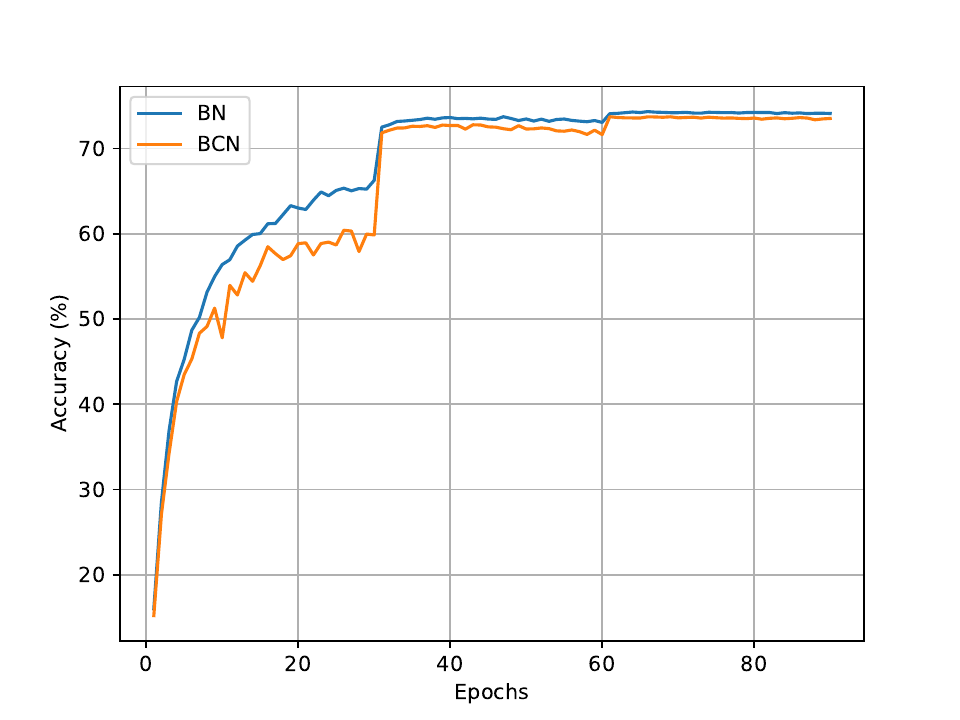}}
      \captionsetup{justification=centering}
    
   \end{subfigure}
   \caption{Testing accuracy of DenseNet with  ImageNet dataset.}
 \label{fig12}
\end{figure}

\subsection{Results on DenseNet}
DenseNet is a typical Dense Convolutional Network. We have implemented DenseNet-201 in PyTorch. A careful selection of learning rate value can lead to better performance results. To this end, we experiment with different learning rates to investigate what suits our datasets and topology. In this paper, we initially set the learning rate to 3e-2. Similarly, we perform experiments to identify a batch size. The batch size 512 is used. Max batch size  is used  to fill in GPU memory when training. In our experiments, we trained for 90 epochs, and every epoch had 2503 iterations. In addition, we record training accuracy by steps and testing accuracy by epochs. We compare the training, testing and validation performance of different normalization techniques by combining DenseNet-201 with BN, and BCN in the ImageNet dataset. As illustrated in Figure \ref{fig11} and Figure \ref{fig12}, the proposed BCN can produce a good  performance.

\subsection{Ablation Study}
In this subsection, we explore the impact of the batch size. We evaluate various batch sizes: 128, 16, and 8. Our findings are shown in Figure \ref{fig5},
indicating that the BCN  yields favorable results with different batch sizes. 
To explore whether BCN mitigates the weakness of BN and LN and, as pointed by \cite{b1}, that BN has a bad performance in the case of small batch size. In this subsection, we focus on addressing the minibatch problem of BN. The experiments show that BCN alleviates the problem of BN with small batch size. BCN achieves good performance with small batch sizes 4 and 2, as shown in Figure~\ref{fig4}. In a batch size of 8 and 20 epochs, BCN has achieved about 86.12\% training accuracy and 84.16\% validation accuracy, whereas in the same number of epochs and 2 batch size, BCN has achieved about 84.58\% training accuracy and 81.42\%  validation accuracy. This proves that the proposed technique works well in a small batch size.


%% file: sec/5_con.tex
\section{Conclusion}
\label{sec:conclusion}
In this paper, we proposed a new normalization technique termed Batch Channel normalization (BCN). It simultaneously exploits the channel and batch dimensions and adaptively combines the normalized outputs. Our experiments on typical models and datasets show that BCN can consistently outperform the state-of-the-art normalization techniques, demonstrating that BCN is a general normalization technique.  An ablation study of directly computing the average and variance along 
(N, C, H, W)  can be done as a future work. Moreover, we will investigate the BCN technique across more applications and evaluate the usefulness of BCN across a wider range of CNN architectures.